\documentclass[
]{ceurart}

\usepackage{listings}
\usepackage{inputenc}
\usepackage[nolist]{acronym} 
\usepackage{subcaption}
\usepackage[disable]{todonotes}
\usepackage{tablefootnote}

\begin{document}

\copyrightyear{2022}
\copyrightclause{Copyright for this paper by its authors.
  Use permitted under Creative Commons License Attribution 4.0
  International (CC BY 4.0).}

\conference{MediaEval'22: Multimedia Evaluation Workshop,
  January 13--15, 2023, Bergen, Norway and Online}
  
\title{Fine-Grained Action Detection with RGB and Pose Information using Two Stream Convolutional Networks}

\author[1]{Leonard Hacker}\cormark[1]%
\author[1]{Finn Bartels}\cormark[1]%
\author[2]{Pierre-Etienne Martin}[%
    orcid=0000-0002-9593-4580,
    email=pierre_etienne_martin@eva.mpg.de,
    url=www.eva.mpg.de/comparative-cultural-psychology/staff/pierre-etienne-martin,
]\fnmark[1]

% Affiliations 
\address[1]{Computer Science Institute, University of Leipzig, Germany}
\address[2]{CCP Department, Max Planck Institute for Evolutionary Anthropology, D-04103 Leipzig, Germany}

\cortext[1]{These authors contributed equally.}
\fntext[1]{Corresponding author.}

\begin{abstract}
As participants of the MediaEval 2022 Sport Task, we propose a two-stream network approach for the classification and detection of table tennis strokes. Each stream is a succession of 3D \ac{CNN} blocks using attention mechanisms. Each stream processes different 4D inputs.
Our method utilizes raw RGB data and pose information computed from MMPose toolbox. The pose information is treated as an image by applying the pose either on a black background or on the original RGB frame it has been computed from. 
Best performance is obtained by feeding raw RGB data to one stream, \ac{PRGB} information to the other stream and applying late fusion on the features.
The approaches were evaluated on the provided TTStroke-21 data sets.
We can report an improvement in stroke classification, reaching 87.3\% of accuracy, while the detection does not outperform the baseline but still reaches an IoU of 0.349 and mAP of 0.110.
\end{abstract}

\maketitle
\vspace{-20pt}
\section{Introduction}
\label{sec:intro}
While there have been great advances in detection of coarse-grained action in videos (e.g. the type of sport being performed), fine-grained action detection is inherently more difficult due to its low inter-class variability~\cite{zhu2020comprehensive,mediaeval/Martin/2022/overview}.
The goal of this benchmark task is to provide viable tools to enable analyzing athletes' performance~\cite{mediaeval/Martin/2022/overview}.
Table Tennis entails many interesting challenges for fine-grained video detection, e.g. ball trajectory prediction~\cite{TT:BallTracking&Trajectory:2020} or real-time score and game analysis~\cite{TT:TTNet:2020}.

\par
In the field of image recognition, a \ac{CNN} consisting of Convolutional layers, ReLU layers and Max Pooling layers is a conventional practice~\cite{under_CNN}.
The provided baseline model in the competition originates from this~\cite{mediaeval/Martin21/baseline,mediaeval/Martin/2022/baseline}.
While video data includes an additional dimension (time or frame number), several approaches adapt a plain \ac{CNN} to determine movement or changes in a range of frames~\cite{zhu2020comprehensive}.
A popular approach introduced first by \citeauthor{simonyvan2014two}~\cite{simonyvan2014two} is the Two-Stream Neural Network.
They implemented an architecture consisting of a spatial and a temporal \ac{CNN}.
The aforementioned spatial stream represents a single frame at each time while the temporal stream is built as a multi-frame \ac{OF} recognizing the momentum of movement in multiple RGB frames.
Hence, the model obtains the additional benefit of \textit{complementary information} provided by a second stream.
Thus, areas within the video covering no movement can be removed easily to reduce noise~\cite{zhu2020comprehensive}. A successful implementation built upon this is the \ac{I3D} model~\cite{carreira2017quo}, where multiple images are pushed into a 3D \ac{CNN} instead of a single image at a time.
\citeauthor{feichtenhofer2019slowfast}~\cite{feichtenhofer2019slowfast} proposed networks called SlowFast based on a two-stream architecture. The first stream processes the frames with a low frame rate and the second stream with a high frame rate so that the semantic information and motion information are considered.  

The fusion step is an essential part of multiple stream networks.
Within the models, there are different approaches differing from where the fusion is performed.
Late fusion is the easiest and one of the most efficient ways to proceed~\cite{PeMEWork:2020,zhu2020comprehensive}.
Fusion can be performed after a fully connected ReLU layer before a final Softmax function~\cite{martin:tel-03099907,PeCBMI:2018}. 
Early fusion is usually performed at an earlier stage of the network, meaning that the information of the second stream is pushed into the first stream~\cite{feichtenhofer2019slowfast,feichtenhofer2016convolutional}.
The following sections describe the architecture of two streams, the results of stroke classification and stroke detection and a discussion on preliminary factors and approaches.

\section{Method}
\label{sec:Method}
As mentioned, the two-stream architecture is one of the state-of-the-art methods.
For this contribution, the baseline provided by MediaEval is extended into a two-stream network utilising raw RGB images and pose information.
The baseline itself is a single stream 3D \ac{CNN} with an attention mechanism.
Recent papers suggest that the utilization of pose information can achieve better results for fine-grained action detection than \ac{OF}~\cite{duan2021revisiting,PeICIP:2019,DBLP:conf/mediaeval/SatoA20}. When considering actions that are performed by people, the pose holds significant information. We superimpose this information by drawing the pose information on top of RGB images and feeding that into a second stream. The implementation of our method is available online\footnote{\url{https://github.com/fidsinn/SportTaskME22}}.

 \begin{figure}
    \centering
    \begin{subfigure}[b]{.325\linewidth}
        \centering
        \includegraphics[width=\textwidth]{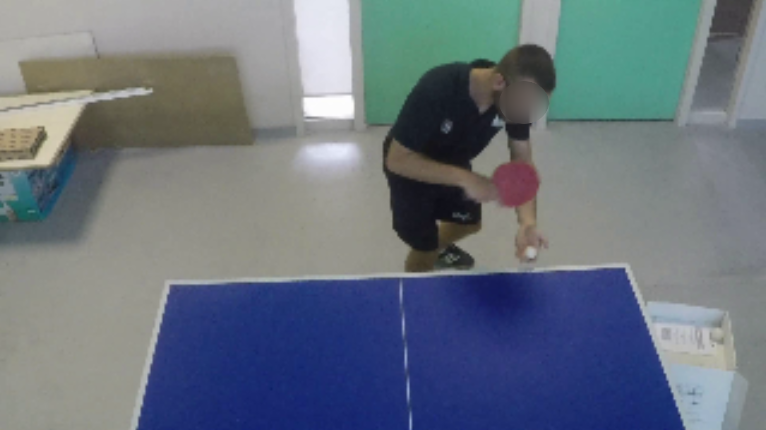}
        \caption{RGB}
        \label{fig:rgb}
    \end{subfigure}
    \hfill
    \begin{subfigure}[b]{.325\linewidth}
        \centering
        \includegraphics[width=\textwidth]{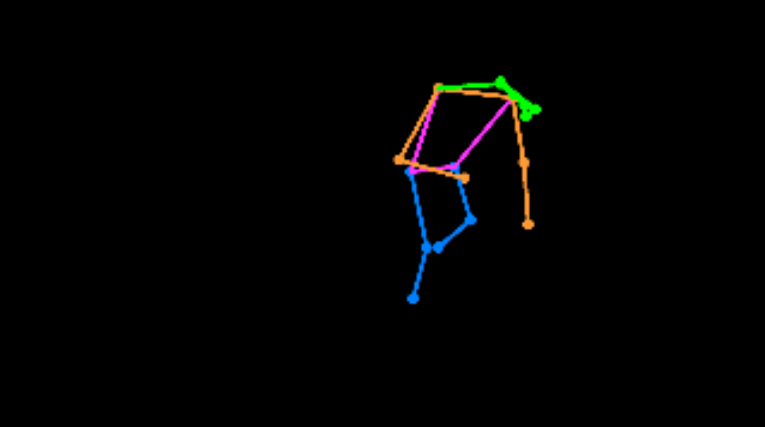}
        \caption{Pose}
        \label{fig:s}
    \end{subfigure}
    \hfill
    \begin{subfigure}[b]{.325\linewidth}
        \centering
        \includegraphics[width=\textwidth]{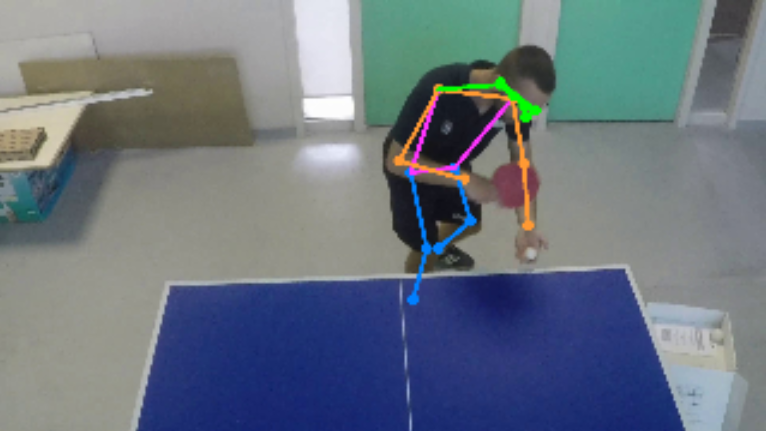}
        \caption{\ac{PRGB}}
        \label{fig:srgb}
    \end{subfigure}
    \caption{RGB, Pose and \ac{PRGB} frames of the TTStroke-21 videos. The generated frames serve as input for the network branches.}
    \label{fig:data}
    \vspace{-10pt}
\end{figure}

\subsection{Pose Estimation}
The pose information is the traced human pose in each frame as depicted in \autoref{fig:data}.
Since \ac{OF} is already well researched and the use of pose information yields promising results, this work focuses on using pose information to the two-stream \ac{CNN} framework.
The pose information is extracted using the MMPose~\cite{mmpose2020} toolbox from OpenMMLab: an open-source package for detailed video understanding.
Each frame of the input video is analyzed.
First, a person detector is used to draw bounding boxes around every person in the frame, then a pose estimator is deployed to extract the poses out of the bounding boxes. For person detection we utilized a faster region-based \ac{CNN}~\cite{ren2015faster} and for pose estimation deep high-resolution representation learning~\cite{sun2019deep}.
A top-down classifier is utilized for keypoint extraction, as it performs better than bottom-up classifiers~\cite{duan2021revisiting,lin2014microsoft}.
Both models were pre-trained on the COCO data set~\cite{lin2014microsoft} by OpenMMLab.
The individual keypoints are then connected by differently colored segments representing body parts superimposed over an image.
Since different body parts contribute to the strokes in different ways, we assume that the model can distinguish them based on the coloring of each body part.
In this work, two methods utilizing pose information are investigated: Pose and \ac{PRGB}.
The Pose variant contains the computed pose over a black background, while the \ac{PRGB} uses the original RGB frame as a background. 
By comparing pose, \ac{PRGB} and RGB performance, it is possible to determine how well the network utilizes the pose information.

\subsection{Architecture}
\begin{figure}
    \centering
    \includegraphics[width=\linewidth]{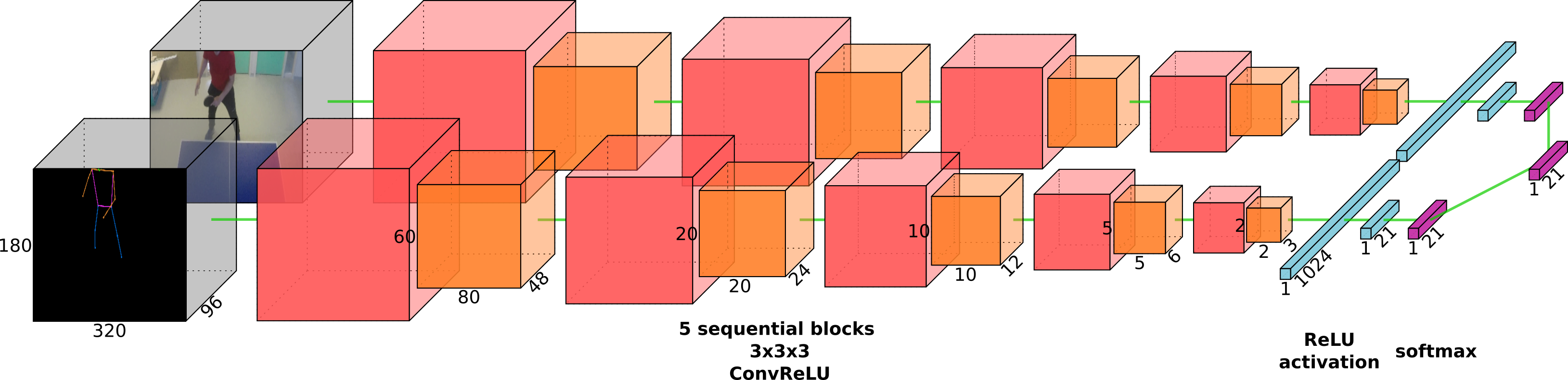}
    \caption{Two Stream Pose Convolutional Neural Network with RGB images (top branch) and Pose information on a black background (lower branch). A branch consists of five successive convolutional layers using ReLU activation function with a growing number of filters (32, 64, 128, 256, 512), each followed by an attention block (both represented in red) and a pooling layer (in orange) of size ($4\times3\times2$) for the two first ones and ($2\times2\times2$) for the others.}
  \label{fig:architecture}
  \vspace{-10pt}
\end{figure}

Our model is a variant of the two-stream architecture~\cite{PeICPR:2020} and an extension of the provided baseline~\cite{mediaeval/Martin/2022/baseline}, which also uses 3D convolutional blocks and an attention mechanism.
As depicted in \autoref{fig:architecture}, the \ac{TSPCNN} consists of two identical streams, each with five convolutional layers and pooling layers with an increasing number of filters leading to a linear layer with ReLU activation. The latest feeds a second linear layer followed by a Softmax function to convert the output into a 21-dimensional probabilistic vector for classification (21 different stroke types including non-stroke class) or a two-dimensional vector for detection (stroke and non-stroke class). The output of the two branches is then summed and processed by the last Softmax function to have a probabilistic output for classification and detection.
The first Softmax function normalizes the output of each individual stream before fusion to minimize vanishing gradients.

\subsection{Fusion}
Literature suggests that employing early fusion combined with late fusion boosts the performance of a two-stream model considerably~\cite{duan2021revisiting}.
Adding multiple fusion methods to the \ac{TSPCNN} showed limited gain in performance. 
The best performance was achieved using a late fusion approach, i.e. fusing before the last layer.
Different fusion styles were i) weighted fusion, where the resulting feature is equal to the weighted sum of the two fused features, ii) summed fusion where the resulting feature is the sum of both features and iii) concatenated fusion, where the resulting feature is a concatenation of both features. Summed fusion performed best, and therefore, only its results are reported in the remaining of this paper.

\subsection{Training}
All experiments were performed on Tesla V100 GPUs provided by the University of Leipzig. The training took 7 to 8 minutes per epoch for the detection task and 1 to 2 minutes for the classification task over 2000 epochs, with a learning rate of 0.0001 and a momentum of 0.5.

\section{Results and Discussion}

We evaluated our approach using the TTStroke-21 data set~\cite{PeMTAP:2020} provided by the Sport task organizers of MediaEval~\cite{mediaeval/Martin/2022/overview}.
The results are compared with the provided baseline~\cite{mediaeval/Martin/2022/baseline}. 
To compare our approach with the baseline, we evaluated our runs using accuracy for the classification task and IoU and mAP for the detection task.
In addition, we added the results for training accuracy and validation accuracy for each model, reported in \autoref{Table:Results}.
The test results were selected depending on which of several decision methods produced the best results~\cite{mediaeval/Martin/2022/baseline}.
The baseline and the other single-stream approaches already achieve quite promising results for the classification task.
Basic RGB combined with PRGB in a two-stream approach shows the best accuracy in testing.
The two-stream approaches slightly improve the classification results compared to the single-stream method by up to .009.
In contrast to the improvement regarding the classification task, for detection the two-stream methods using the pose stream lead to a decreasing quality compared to the single-stream baseline.
The poor detection performance is likely due to the missing ball and racket information in the pose data.
An arm movement without the racket may also be similar to a stroke which can confuse the model.
Therefore, we can say that pose information did not improve stroke detection performance.

\begin{table}
    \caption{Results on classification accuracy (\%) and detection metrics (accuracy, IoU, mAP (\%)) of baseline, one stream approaches and two stream approaches.}
    \label{Table:Results}
    \centering
        \begin{tabular}{|c|ccc|cccc|}
        \hline
        \multirow{2}{*}{Models} & \multicolumn{3}{c|}{Classification}                                 & \multicolumn{4}{c|}{Detection}                                                                  \\ \cline{2-8} 
                                & \multicolumn{1}{c|}{Train} & \multicolumn{1}{c|}{Validation} & Test & \multicolumn{1}{c|}{Train} & \multicolumn{1}{c|}{Validation} & \multicolumn{1}{c|}{Test IoU}  & Test mAP  \\ \hline
        Baseline                & \multicolumn{1}{c|}{-}     & \multicolumn{1}{c|}{0.813}       & 0.864$^2$ & \multicolumn{1}{c|}{-}     & \multicolumn{1}{c|}{-}          & \multicolumn{1}{c|}{\textbf{0.515}$^5$ (0.365$^3$)} & \textbf{0.131}$^5$ (0.118$^4$) \\ \hline
        Pose                    & \multicolumn{1}{c|}{0.995} & \multicolumn{1}{c|}{\textbf{0.878}}       & 0.847$^1$ & \multicolumn{1}{c|}{0.862}  & \multicolumn{1}{c|}{0.591}       & \multicolumn{1}{c|}{0.205$^1$} & 0.046$^1$ \\ \hline
        PRGB                    & \multicolumn{1}{c|}{0.978}  & \multicolumn{1}{c|}{0.813}       & 0.864$^1$ & \multicolumn{1}{c|}{0.980}  & \multicolumn{1}{c|}{0.834}       & \multicolumn{1}{c|}{0.165$^1$} & 0.036$^1$ \\ \hline
        RGB and Pose            & \multicolumn{1}{c|}{1}     & \multicolumn{1}{c|}{0.830}       & 0.872$^1$ & \multicolumn{1}{c|}{0.987}  & \multicolumn{1}{c|}{0.820}       & \multicolumn{1}{c|}{0.331$^1$} & 0.100$^1$ \\ \hline
        RGB and PRGB            & \multicolumn{1}{c|}{0.998}  & \multicolumn{1}{c|}{0.848}       & \textbf{0.873}$^1$ & \multicolumn{1}{c|}{0.990}  & \multicolumn{1}{c|}{\textbf{0.840}}       & \multicolumn{1}{c|}{0.349$^1$} & 0.110$^1$ \\ \hline
        \end{tabular} 
    \footnotesize{Decision method: $^1$No Window, $^2$ Gaussian, $^3$ Mean, $^4$ Vote, $^5$ Vote (Sliding Window)}
\end{table}

%\begin{figure}
%    \begin{tabular}{cc}
%        \includegraphics[width=.45\linewidth]{images/cm_serve.png} &
%        \includegraphics[width=.45\linewidth]{images/cm_hand.png} \\
%        \textbf{a.} Type & \textbf{b.} Hand-side 
%    \end{tabular}
%    \caption{Confusion matrices with higher level categories.}
%    \label{fig:confless}
%\end{figure}
We have shown that the pose information can be suited for fine-grained action classification but seems to fail to capture discriminant features for detection.
While the \ac{TSPCNN} outperformed the baseline in the classification task, we could not improve detection performance.
A contributing factor to the slight classification improvement might be the limited training data, especially since some classes only have a few labelled videos in the training set.

\section{Summary and Outlook}

As wearable systems can be intrusive and cumbersome to set up while also not being widely available, fine-grained action detection from video is of high interest for athletes and coaches to be able to classify different actions in their game and to improve training efficiency~\cite{mediaeval/Martin/2022/overview}.
We have built a \ac{TSPCNN} on state-of-the-art research. Our main contribution is to use RGB data overlaid with human poses in a two-stream network.
Our approach slightly outperforms the baseline in terms of classification accuracy but produces poor performances for stroke detection.
To improve the \ac{TSPCNN} further, some more experiments with different qualities of pose data are needed.
Moreover, different representations of pose data can be evaluated such as thicker lines to emphasise poses.
The approach should also be validated with different data sets, such as the Finegym data set~\cite{Dataset:Gym:2020} since it also has low variability between classes but a more even class distribution in the training data.

\begin{acronym}[CNN]
\acro{R-CNN}{region-based Convolutional Neural Network}
\acro{HRRL}{High-Resolution Representation Learning}
\acro{CNN}{Convolutional Neural Network}
\acro{I3D}{Inflated 3D Convolutional Neural Network}
\acro{PRGB}{Pose + RGB}
\acro{TSPCNN}{Two Stream Pose Convolutional Neural Network}
\acro{OF}{Optical Flow}
\end{acronym}

\def\bibfont{\small} % comment this line for a smaller fontsize
\bibliography{references} 

\end{document}